%% file: main.tex
\documentclass[11pt]{article}
\usepackage[top=1in, bottom=1in, left=1in, right=1in, letterpaper]{geometry}

\usepackage[T1]{fontenc}

\usepackage[sc,osf]{mathpazo}   
\usepackage[euler-digits,small]{eulervm}

\usepackage{booktabs}       
\usepackage{amsmath, amssymb, amsfonts, bm}
\usepackage{amsthm}
\usepackage{mathtools}
\usepackage{nicefrac}       
\usepackage{microtype}      
\usepackage{subcaption}
\usepackage[usenames, dvipsnames]{color}
\usepackage{floatrow}
\usepackage{graphicx}
\usepackage{algorithm}
\usepackage{algorithmicx}
\usepackage[noend]{algpseudocode}
\usepackage{amssymb}
\usepackage{multirow,bigdelim}
\usepackage{textcomp}
\usepackage[title]{appendix}
\usepackage[percent]{overpic}
\usepackage{floatrow}
\usepackage{textcomp}
\usepackage{xcolor}

\def\BibTeX{{\rm B\kern-.05em{\sc i\kern-.025em b}\kern-.08em
    T\kern-.1667em\lower.7ex\hbox{E}\kern-.125emX}}

\usepackage{subcaption}
\usepackage{booktabs} 
\usepackage{algorithm}
\usepackage{algorithmicx}

\usepackage{cite}
\usepackage[hyphens]{url}
\usepackage[colorlinks=true,citecolor=blue,linkcolor=blue,bookmarksnumbered=true,hyperfootnotes=true,breaklinks]{hyperref}
\usepackage[nameinlink]{cleveref}

\newcommand{\Tab}[1]{Table.~\ref{tab:#1}}
\newcommand{\fig}[1]{Fig.~\ref{fig:#1}}
\newcommand{\Fig}[1]{Fig.~\ref{fig:#1}}

\DeclareMathOperator{\ctrl}{control}
\DeclareMathOperator{\sharpen}{sharpen}
\DeclareMathOperator{\conv}{conv}

\begin{document}

\title{Using Multi-task and Transfer Learning to \\ Solve Working Memory Tasks}

\author{%
	T.S. Jayram 
	\and Tomasz Kornuta 
	\and Ryan L. McAvoy 
	\and Ahmet S. Ozcan \\
	IBM Research AI, Almaden Research Center, San Jose, USA\thanks{%
		Contacts: \texttt{\{jayram, tkornut, mcavoy, asozcan\}@us.ibm.com}}
}
\date{}

\maketitle

\begin{abstract}
We propose a new architecture called \emph{Memory-Augmented Encoder-Solver (MAES)} that enables transfer learning to solve complex working memory tasks adapted from cognitive psychology. It uses dual recurrent neural network controllers, inside the encoder and solver, respectively, that interface with a shared memory module and is completely differentiable. We study different types of encoders in a systematic manner and demonstrate a unique advantage of  multi-task learning in obtaining the best possible encoder. We show by extensive experimentation that the trained MAES models achieve task-size generalization, i.e., they are capable of handling sequential inputs 50 times longer than seen during training, with appropriately large memory modules. We demonstrate that the performance achieved by MAES far outperforms existing and well-known models such as the LSTM, NTM and DNC on the entire suite of tasks.
\end{abstract}

%

\input{intro}

\input{tasks_new}

\input{architecture}

\input{experiments}

\input{summary}

\bibliographystyle{abbrv}
\bibliography{sample}

\end{document}

%% file: intro.tex
\section{Introduction}
\label{sec:intro}
Recent progress in the field of Deep Learning~\cite{lecun2015deep} introduced several crucial improvements to neural networks, such as gating mechanisms~\cite{hochreiter1997long,chung2014empirical} and attention~\cite{bahdanau2014neural,weston2015memory,vaswani2017attention}.
It also reinvigorated the interest in augmenting neural networks with external memory modules to extend their capabilities in solving diverse tasks, e.g. 
learning context-free grammars~\cite{joulin2015inferring},
remembering long sequences (long-term dependencies)~\cite{gulcehre2017memory},
learning to rapidly assimilate new data (e.g. one-shot learning)~\cite{santoro2016meta} 
and visual question answering~\cite{ma2017visual}. In addition they showed promise in algorithmic tasks such as copying sequences, sorting digits~\cite{graves14} and traversing graphs~\cite{graves2016hybrid}.
Some of the mentioned studies are inspired by the human memory and make links to working~\cite{graves14} or episodic memory~\cite{graves2016hybrid}, however, the tasks they solve are not necessarily limited to the type of memory they model. This is probably due to the desire to build a general memory architecture, which is capable of handling a multitude of tasks. 
The algorithmic applications of such models are particularly interesting, since they provide better opportunities to analyze the capabilities, generalization performance and the limitations of those models.

Working memory in the brain refers to the ability to keep information in mind after it is no longer present in the environment,
and is distinct from long-term memory which is responsible for the storing vast amount of information~\cite{baddeley2003}.  
There is a rich literature in cognitive psychology on \emph{working memory tasks} that shed light into the properties and
underlying mechanisms~\cite{wilhelm2013, conway2005, unsworth2005, shipstead2012}. 
In solving such tasks, we identified the following four core requirements for working memory:
1) encoding the input information into a useful representation in memory;
2) retention of information during processing, i.e.,  to prevent interference and corruption of the memory content;
3) using the encoding to solve the task, i.e., producing a suitable output; and
4) controlled attention in memory during encoding, processing and solving.

In this paper, we propose a novel architecture called \emph{Memory-Augmented Encoder-Solver (MAES)}
based on the above requirements of working memory.
The architecture uses a dual controller neural network design, representing the encoder and solver, respectively,
with a shared memory module and is completely differentiable.
We study different types of encoders in systematic manner and demonstrate a unique advantage of 
\emph{multi-task learning}~\cite{caruana1993multitask} in obtaining the best possible encoder.
We demonstrate that this encoder enables \emph{transfer learning}~\cite{pan2010survey}  
to solve a suite of working memory tasks.
To our knowledge, this is the first demonstration of transfer learning for neural networks with external memory, 
in contrast to prior works that showed tasks learned in separation.
Finally, we show by extensive experimentation that the trained models achieve \emph{task-size generalization}, i.e.,
they are capable of dealing with input sequences 50 times longer than the ones seen during training,
with appropriately large memory modules. We demonstrate that the performance achieved by MAES
far outperforms existing models including the classical LSTM (Long Short-Term Memory)~\cite{hochreiter1997long} as well as
NTM (Neural Turing Machine)~\cite{graves14} and DNC (Differentiable Neural Computer)~\cite{graves2016hybrid}---two seminal differentiable models hat use external memory.

%% file: tasks_new.tex
\section{Working Memory Tasks}
\label{sec:tasks}

\begin{figure*}[htbp]
    \centering
    \begin{subfigure}[t]{0.28\textwidth}
		\includegraphics[width=\textwidth]{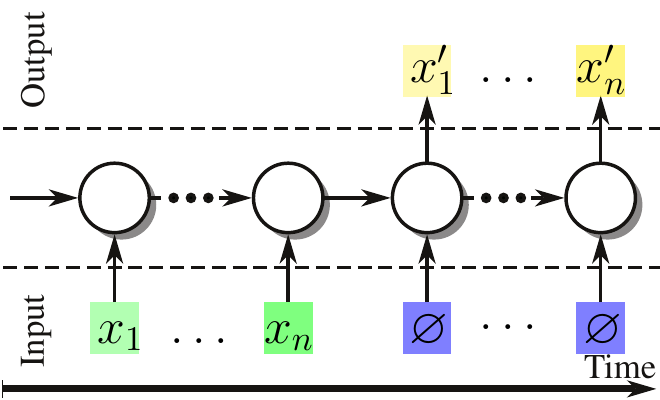}
        \caption{Serial Recall}
		\label{fig:serial_recall}
    \end{subfigure}
    \hfill
    \begin{subfigure}[t]{0.28\textwidth}
		\includegraphics[width=\textwidth]{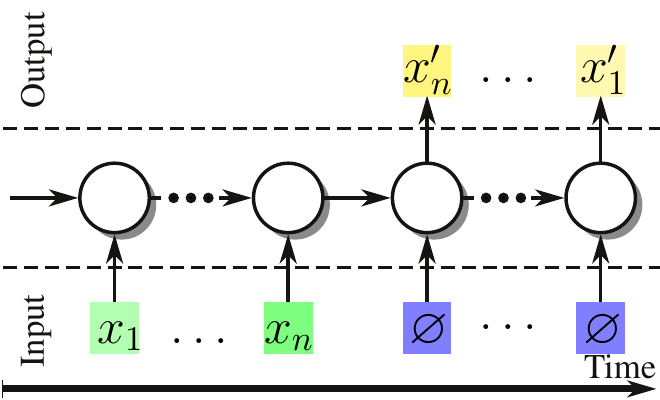}
        \caption{Reverse Recall}
		\label{fig:reverse_recall}
    \end{subfigure}
    \hfill
    \begin{subfigure}[t]{0.40\textwidth}
		\includegraphics[width=\textwidth]{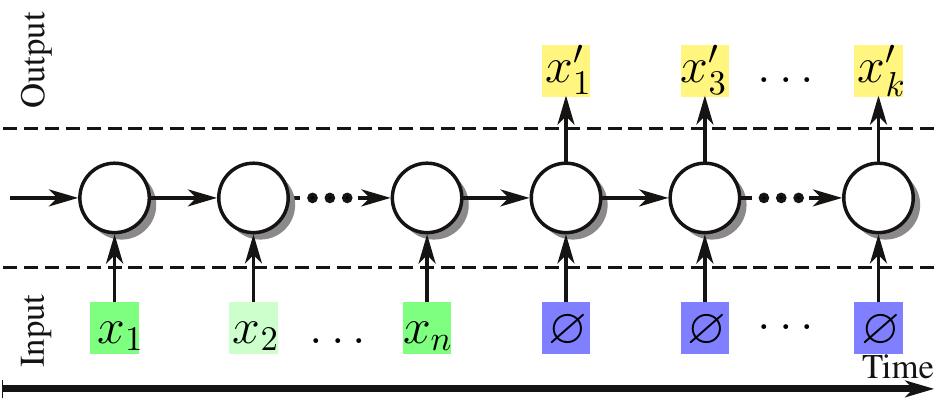}
		\caption{Odd Recall}
		\label{fig:odd_recall}
    \end{subfigure}
       
    \begin{subfigure}[t]{0.28\textwidth}
		\includegraphics[width=\textwidth]{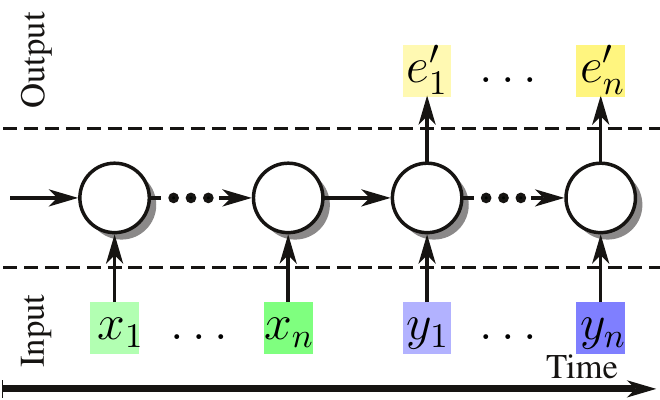}
		\caption{Sequence Comparison}
		\label{fig:sequence_comparison}
    \end{subfigure}
    \quad\quad
    \begin{subfigure}[t]{0.28\textwidth}
		\includegraphics[width=\textwidth]{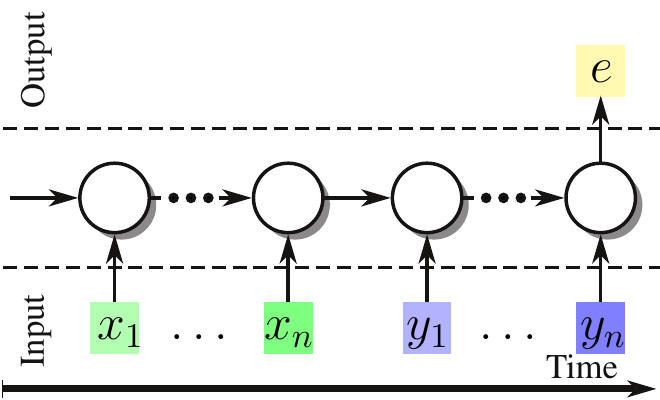}
		\caption{Sequence Equality}
		\label{fig:sequence_equality}
    \end{subfigure}
       
    \caption{A selected suite of working memory tasks}
	\label{fig:tasks}
\end{figure*}

\Fig{tasks} presents the five tasks adapted from cognitive psychology that we consider in this paper.  
The first three tasks, namely \emph{Serial Recall}, \emph{Reverse Recall} and \emph{Odd Recall},  
involve simple (or no) manipulation of the input sequence. 
Crucially, to emphasize encoding and retention of input in memory, the output is required to be given 
only after the entire input sequence has been read by the model. 
While these tasks are relatively simple, they are important in designing good encoders using multi-task learning
that are used for solving the three remaining complex tasks via transfer learning.
These involve relational manipulation using a main and an auxiliary sequence.
Here the output is produced only after the main input has been fully read and (concurrently) while the 
auxiliary sequence is processed.
In \emph{Sequence Comparison}, the relational manipulation involves comparing (for equality) 
corresponding elements of the main and auxiliary sequence in order.
The final task \emph{Sequence Equality} seems significantly more difficult because the predicate that defines it
requires knowing all element-wise comparisons in order to compute the answer.
As the supervisory signal provides only one bit of information at the end of two sequences with varying length, there is an extreme disproportion between the information content of input and output data, making the task quite challenging.

%% file: architecture.tex
\section{Architecture of a Memory-Augmented Encoder-Solver}
\label{sec:architecture}

\begin{figure}[!b]
    \centering
		\centerline{\includegraphics[width=0.7\columnwidth]{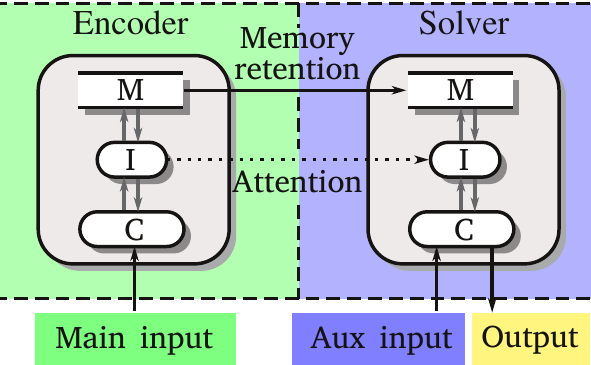}}
		\caption{Memory-Augmented Encoder-Solver (MAES)}
		\label{fig:maes}
\end{figure}

\Fig{maes} shows the basic architecture of a Memory-Augmented Encoder-Solver (MAES). 
It contains neural-network based \emph{dual controllers} that implement the encoder
and solver, respectively.
Both of them interface with a shared \emph{external memory} module represented
by a two-dimensional array $M \times N$ of real numbers.
Here $N$ is the number of addresses and $M$ denotes content at address (analogous to the word size).
The encoder's role is to read the main input and store it in the memory in a suitable form;
it does not produce any output by itself. 
To do that it generates suitable parameters that are used by the interface module to 
generate the address to be accessed and optionally the data to update the memory.
To keep the model differentiable, the addressing is implemented using soft attention, i.e.,
a $N$-dimensional vector with non-negative components summing to 1.
Due to the nature of tasks considered in this paper, it was sufficient to use a single attention mechanism for each 
of the encoder and decoder.
At the end of encoding, the memory contents and, optionally, the attention vector are  
used to initialize the solver; this in effect represents the sharing of memory between the
encoder and solver.
Using the encoding present in memory, as well as the auxiliary input, the solver attempts to complete the task  
by producing the desired output.

\begin{algorithm}[!b]
	\caption{Operation of Encoder/Solver during time step $t$}
	\label{algo:enc}
	\begin{algorithmic}[1]
			\State{$r_t \gets M_{t-1} w_{t-1}$} \Comment{Read from memory}
			\State{$h_t, y_t, P_t \gets \ctrl(x_t, r_t, h_{t-1})$} \Comment{$P_t = (a_t, e_t, s_t, \gamma_t)$}
			\State{$M_t = M_{t-1} \circ (1-e_t) w_t^T + a_t w_t^T$} \Comment{Memory update}
			\State{$w_t = \sharpen(\conv(w_{t-1}, s_t), \gamma_t)$} \Comment{Attention update}
	\end{algorithmic}
\end{algorithm}

\Cref{algo:enc} shows the generic operation of encoder/solver during  a single time step $t$.
At the beginning of time step $t$, $h_{t-1}$ is the hidden state of the recurrent neural network controller,  
$M_{t-1}$ is the $N \times M$ array representing the memory contents, and $w_t$ is the $N $-dimensional
attention vector.
To prevent the hidden state from acting as a separate working memory, we chose its size to be smaller than that of a single input vector. 
First, the memory is accessed through soft attention to produce an $M$-dimensional read vector $r_t$.
This is fed to the controller neural network along with the previous hidden state $h_{t-1}$ and the input $x_t$.
To have finer control, we  handle the controller outputs using separate neural networks.
For the hidden state, it suffices to use a single-layer recurrent neural network controller:
$h_t = \sigma(W_h[x_t, h_{t−1}, r_t])$.
Next, we consider the neural network corresponding to output $y_t$, which is needed only for the solver.
In our design we treat the output as logits and use a simple feedforward neural network with activations in hidden layers.
Last, we compute the interface parameters: $P_t =  (a_t, e_t, s_t, \gamma_t) = W_P[x_t,h_{t−1},r_t]$.
The parameters $e_t$ and $a_t$ correspond to erase and add vectors of dimension $M$;
the memory update is similar to the cell state update in LSTM~\cite{hochreiter1997long} or the 
erase/add step in NTM~\cite{graves14}.
The other two parameters $s_t$ and $\gamma_t$ are used for shifting the attention over memory sequentially;
we use convolution followed by a sharpening step similar to the one used by an NTM.

%% file: experiments.tex
\section{Experiments}
\label{sec:experiments}

\subsection{Methodology}
Throughout the experiments, we fixed the input item size to be 8 bits, so that our sequences
consist of 8-bit words of arbitrary length.
To provide a fair comparison of the training, validation, and testing for the various tasks,
we fixed the following parameters for all the MAES models:
\begin{enumerate}
\item the real vectors stored at each memory address are 10-dimensional, 
and sufficient to hold one input word.
\item the recurrent controllers of both encoders and solvers possessed 5 hidden units.
\item the solvers' output neural network varied from one task to another but the largest that we needed
was a 2-layer feedforward network with a hidden layer of 10 units.
This was necessary for tasks such as sequence comparison and equality, where we need to
perform element-wise comparison on 8-bit inputs (this is closely related to the XOR problem). 
For the other tasks, again a one layer network was sufficient.
\end{enumerate}
The largest network that we had to train still contained less than 2000 trainable parameters.
Note that in MAES models the number of trainable parameters does not depend on the size of the memory.
However, we do have to fix the size of the memory if we want the various parts of MAES, such as the memory or the soft attention of read and write heads, to have a bounded description. 
Thus, we can think of MAES as representing a class of RNNs 
where each RNN is parameterized by the size of the memory, and 
each RNN in principle can take arbitrarily long sequences as its input. 
During training, we fix one such memory size and train with sequences that are ``short enough''
for that memory size; this yields a particular fixing of the trainable parameters.
But now since the MAES model can be instantiated for any choice of memory size, for longer
sequences we can always pick an RNN from a different class corresponding to a larger memory
size and ask whether that RNN (with the same trained parameters)
can solve the task for the same sequence.

In our training experiments, the memory size was limited to $N=30$ addresses,
and we chose sequences of random lengths between 3 and 20. 
The sequence itself also consisted of randomly chosen 8-bit words. This ensured that the input 
data did not contain any fixed patterns so that the trained model doesn't memorize the patterns
and can truly learn the task across all data.
The (average) binary cross-entropy was used as the natural loss function to minimize during training since all of
our tasks, including the tasks with multiple outputs, involved the atomic operation of
comparing the predicted output to the target in a bit-by-bit fashion.
It turned out that for all the tasks, except sequence comparison and equality, the batch size 
did not affect the training performance significantly
so we fixed the batch size to be 1 for all these tasks.
For equality and sequence comparison we chose a batch size of 64 (see Section~\ref{sec:seq-eq} for more details).

During training, we also periodically performed validation on a batch of 64 random 
sequences, each of length 64. The memory size was increased to 80 so that the encoding 
could still fit into memory. This already is a mild form of task-size generalization. 
For all the tasks, we observed that as soon as the loss function dropped to 0.01 or less, 
the validation accuracy was at 100\%. 
However this did not necessarily result in perfect accuracy while measuring task-size generalization 
for much larger sequence lengths. 
To ensure that this would happen, the training was continued until the loss function value was
$10^{-5}$ or less for all the tasks.
The key metric was the number of iterations required to reach this loss value, 
if at all possible, at which point we declared the training to have \emph{(strongly) converged}.
Our data generators could produce an infinite number of samples so we could, in principle,
be training forever. But in cases where we succeeded, the convergence would happen within
20,000 iterations, hence, we stopped the training only if it did not converge in 100,000 iterations.

To measure true task-size generalization, we tested the network on sequences
of length 1000 which required a larger memory module of size 1024.
Since the resulting RNNs were quite huge in size, we performed testing on smaller batch
size of 32 but then averaged over 100 such batches containing random sequences.

\subsection{Transfer learning of encoder pretrained on Serial Recall}

\begin{figure}[htbp]
\centerline{\includegraphics[width=0.7\columnwidth]{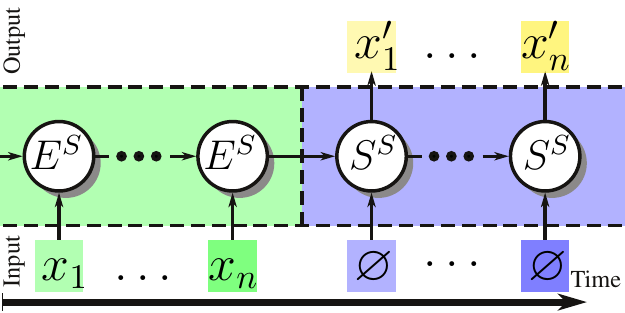}}
\caption{MAES model trained on Serial Recall task in an end-to-end manner}
\label{fig:recall_esr_ssr}
\end{figure}

At first we composed our MAES model, as presented in \fig{recall_esr_ssr}, 
and trained it on the Serial Recall task in an end-to-end manner.
In this simple setting, the goal of the encoder $E^S$ (from Encoder-Serial) was to encode and store the inputs in memory, 
whereas the goal of the solver $S^S$ (from Solver-Serial) was simply to reproduce the output.

\begin{figure}[htbp]
\centerline{\includegraphics[width=0.9\columnwidth]{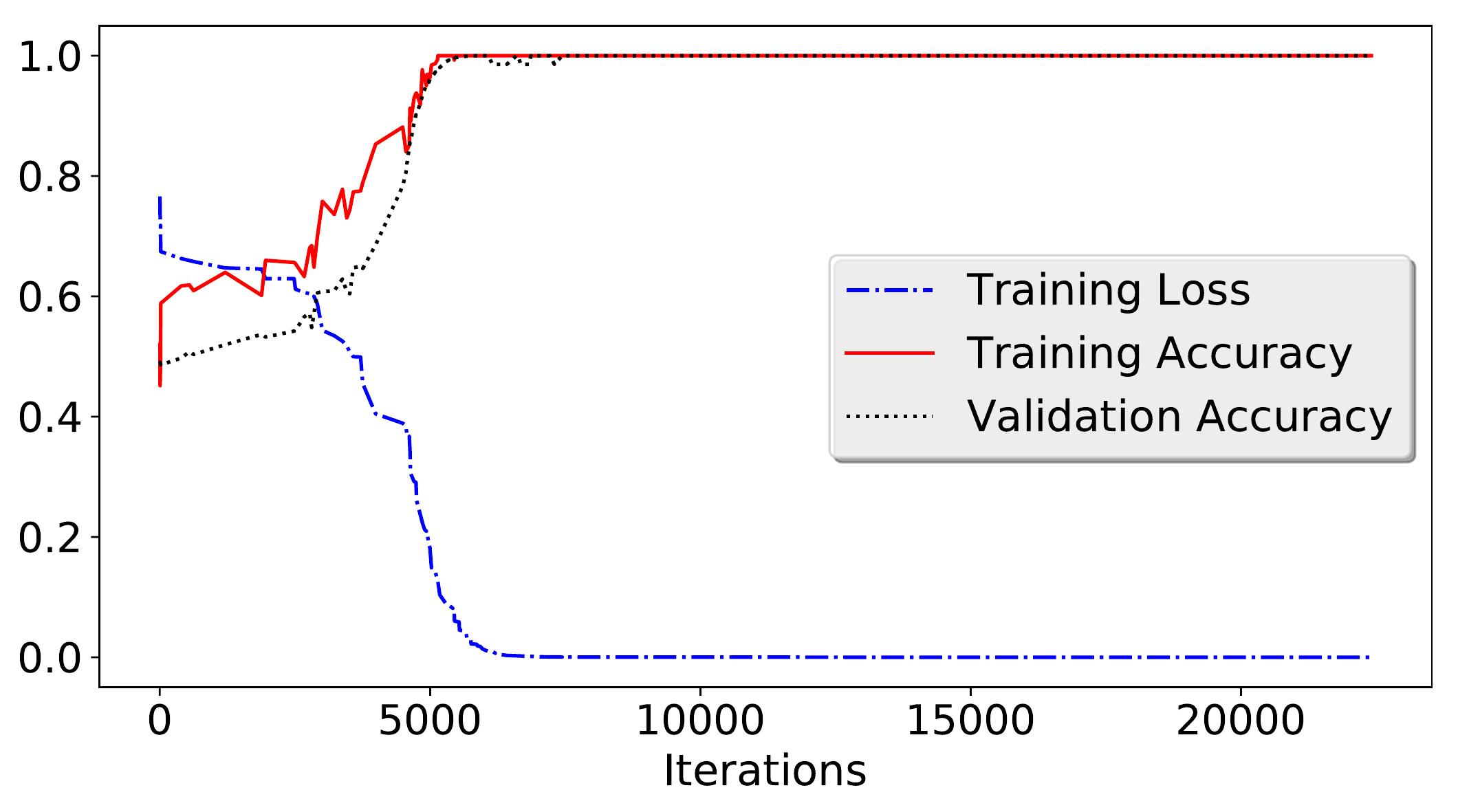}}
\caption{Training performance of MAES trained on Serial Recall task in end-to-end manner.
Training loss of $10^{-5}$ obtained after \texttildelow{}11,000 iterations
}
\label{fig:training_esr_ssr}
\end{figure}

\Fig{training_esr_ssr} shows the training performance with this encoder design.
This procedure took about 11,000 iterations for the training to converge while achieving
perfect accuracy for task-size generalization on sequences of length 1000.

\begin{figure}[!h]
\centerline{\includegraphics[width=0.7\columnwidth]{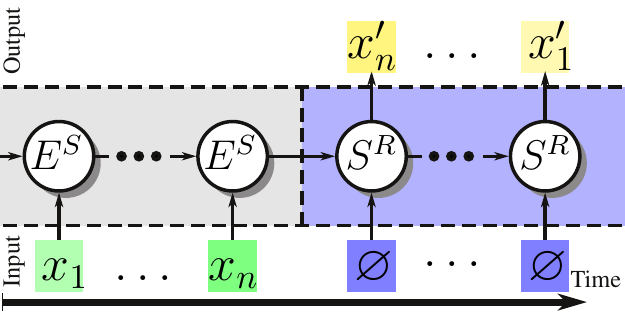}}
\caption{MAES model used for Reverse Recall task. Grey background indicates the frozen part of the model -- we used encoder $E^S$ pretrained on the Serial Recall task ($S^R$ stands for Solver-Reverse)}
\label{fig:recall_esr_fix_srr}
\end{figure}

In the next step we  attempted to reuse the trained encoder $E^S$ for other tasks.
For that purpose we used transfer learning and connected the pretrained $E^S$ with frozen weights with 
new, freshly initialized solvers.
Exemplary MAES model for Reverse Recall task is presented in \fig{recall_esr_fix_srr}.

\begin{table}[!b]
\vskip 0.15in
\begin{center}
\begin{small}
\begin{sc}
\begin{tabular}{lcc}
\toprule
\multirow{2}{*}{Task}& Steps to & Task-size \\
 & converge & generalization \\
\midrule
Serial Recall & 6,000 & $100\%$ \\
Reverse Recall & Fail & Fail \\
Odd & 6,900 & $100\%$ \\
Equality & 27,000 & $100\%$ \\
Comparison & 100,000 & $99.4\%$ \\
\bottomrule
\end{tabular}
\end{sc}
\end{small}
\end{center}
\vskip -0.1in
\caption{Results for MAES using encoder $E^S$ pretrained on Serial Recall task}
\label{tab:MAES-esr-results}
\end{table}

The results are presented in Table~\ref{tab:MAES-esr-results}.
Clearly the training time is reduced by nearly half even for the serial recall which was used to pretrain the encoder.
Moreover, this was good enough to handle the forward processing sequential tasks such as 
Odd and Equality. For sequence comparison, the training did not converge and the 
loss function value could only get as small as $0.02$ but, nevertheless, task-size generalization was about 99.4\%.
For the Reverse Recall task, the training failed completely and the validation accuracy did no better
than random guessing.

\begin{figure}[!t]
    \centering
     \includegraphics[width=\columnwidth]{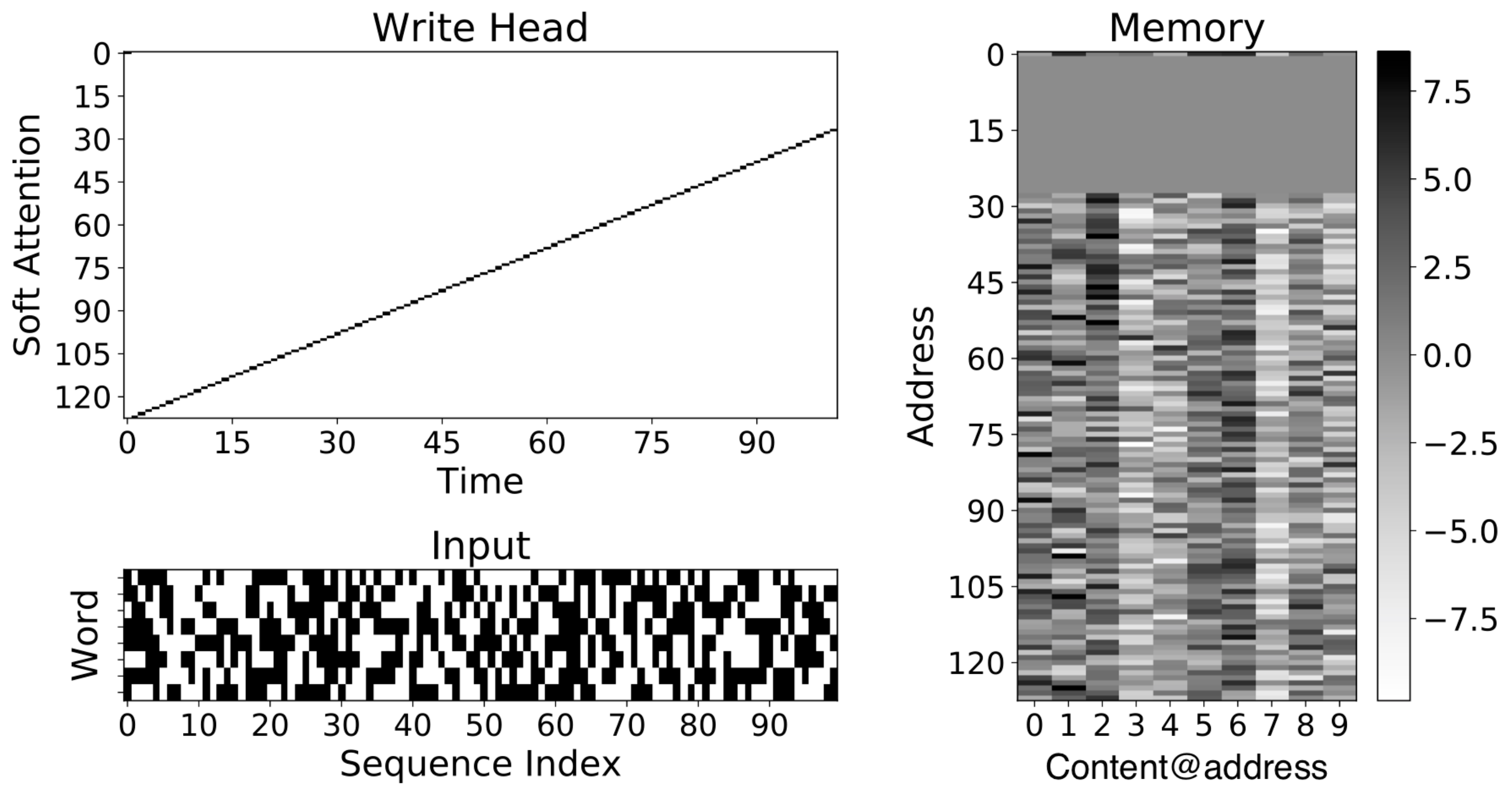}
    \caption{$E^S$ encoder's write attention during processing and memory map at the end for the given input}
     \label{fig:basic-enc-map}
\end{figure}

\begin{figure}[!b]
    \null\hfill
    \begin{subfigure}[t]{0.48\columnwidth}
    	\centering
        \includegraphics[width=\textwidth]{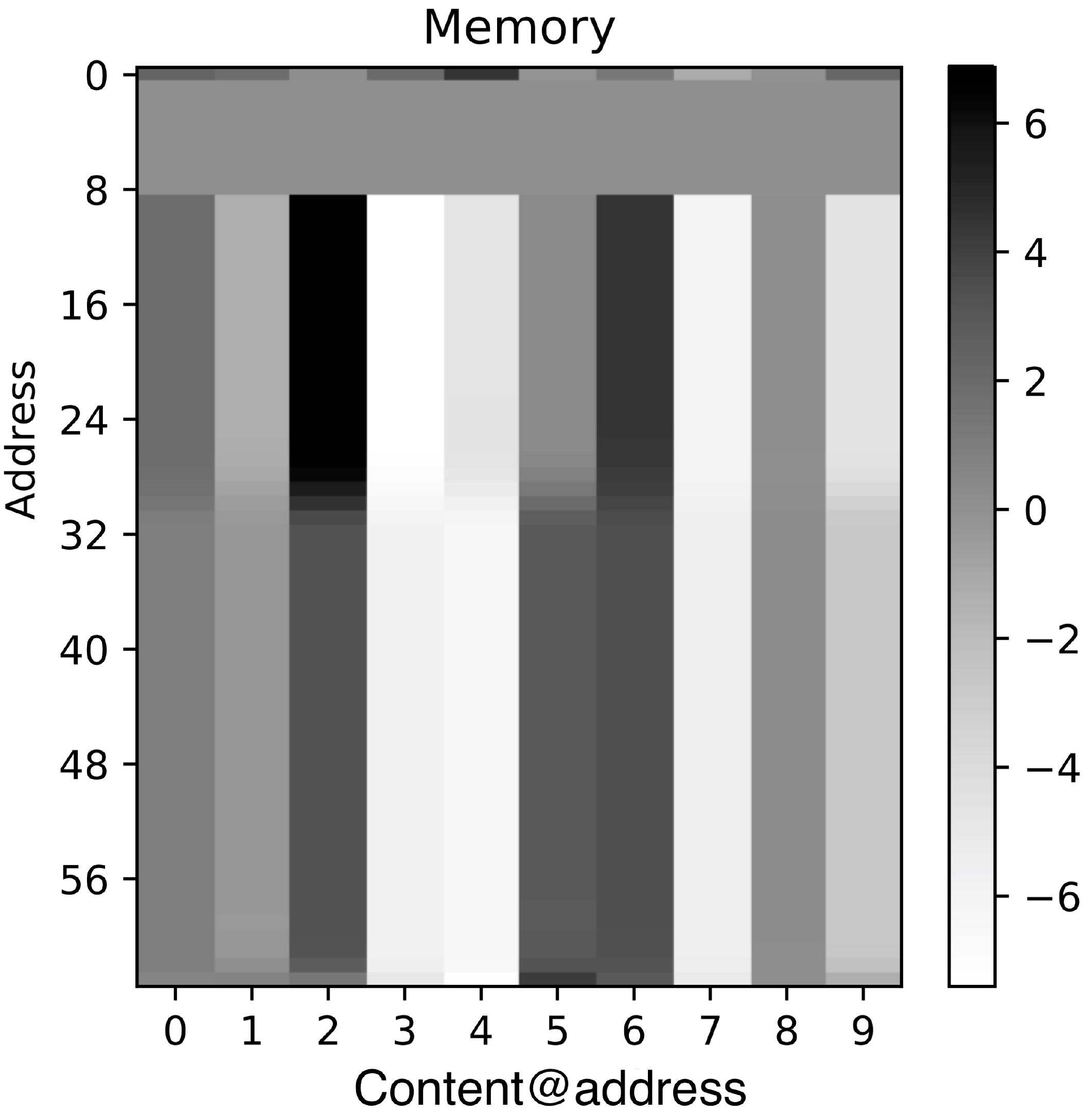}
        \caption{Using the encoder $E^S$}
        \label{fig:basic-enc-same-input}
    \end{subfigure}
    \hfill
    \begin{subfigure}[t]{0.48\columnwidth}
    	\centering
        \includegraphics[width=\textwidth]{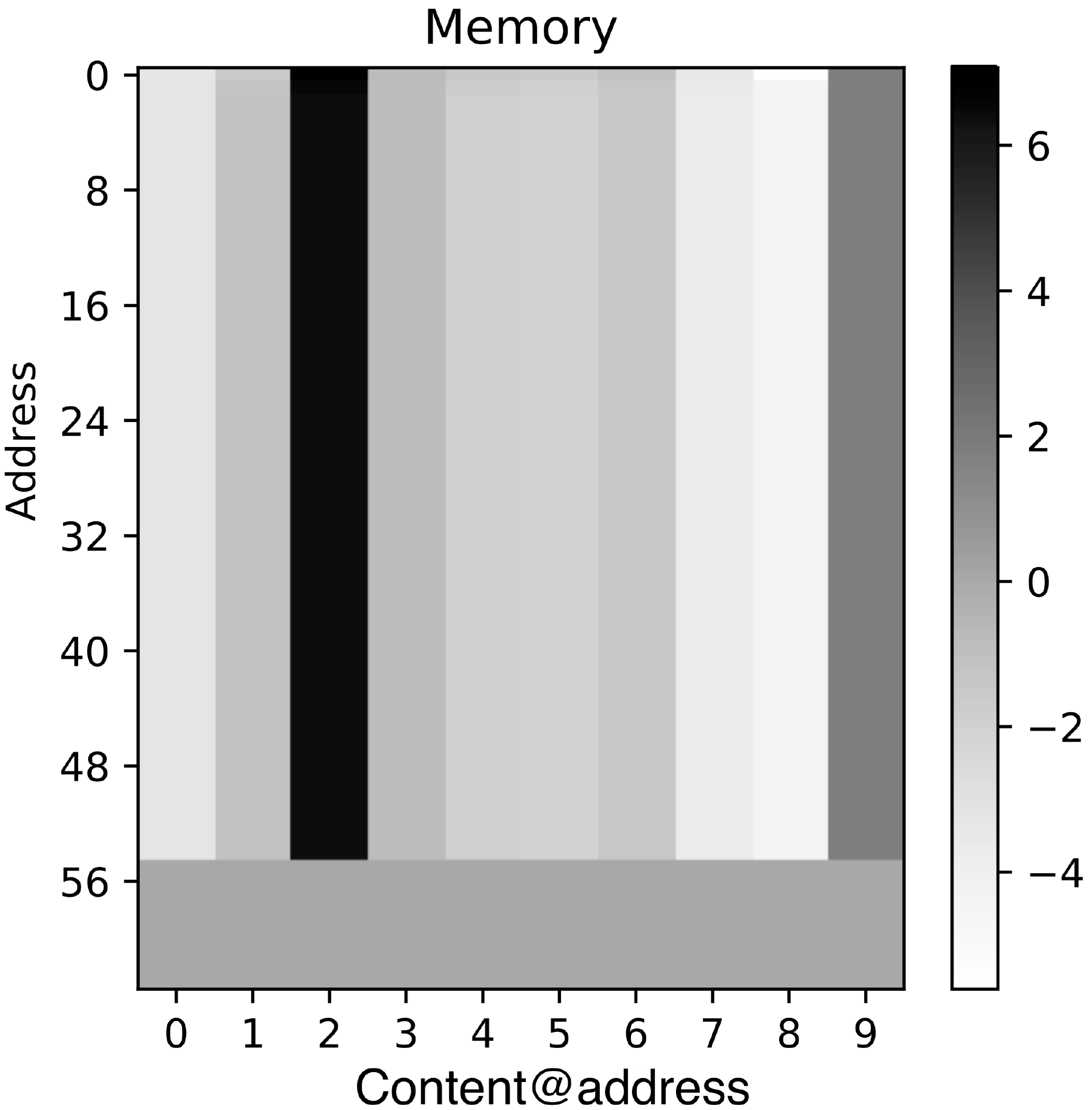}
        \caption{Using the encoder $E^J$}
        \label{fig:gp-enc-same-input}
    \end{subfigure}
    \hfill\null
    \caption{Memory contents after storing a sequence consisting of the same repeated element (right content is the desired one)}
\label{fig:enc-same-input}
\end{figure}

To investigate this further, we designed two experiments to study the behavior of the $E^S$ encoder.
The goal of the first experiment was to validate whether each input is encoded and stored under exactly one memory address.
\Fig{basic-enc-map} shows the write attention as a randomly chosen input sequence of length 100
is being processed and the memory has 128 addresses.   
As can be seen, the trained model essentially uses only hard attention to write to memory. Furthermore, each  write operation is applied to a different location in the memory and these occur sequentially. 
This was universally observed for all the encoders that we tried under different choice of the random seed initialization. 
The only difference was that in some cases it used the lower portion of the memory while in this case the upper portion of the memory addresses was used -- this results from the fact that in some cases (i.e. separate training episodes) the encoder has learned to shift the head one address forward, and in the other backward. 
Thus, we can infer that the encoding of the $k$-th element is $k-1$ locations away from the location where the first element is encoded (viewing memory addresses in a circular fashion).

In the second experiment we looked at what happens when the encoder is fed a sequence consisting of the same element being repeated throughout -- in such a task it would be desired that the content of every memory address where the encoder decided to write should be exactly the same (as seen in \fig{gp-enc-same-input} for an encoder to be described later).
However, as presented in \Fig{basic-enc-same-input}, when the encoder $E^S$ is operational, not all locations are encoded in the same manner and there are slight variations between the memory locations.
This indicated that the encoding of each element is also influenced by the previous elements of 
the sequence as well.
In other words, the encoding  has some sort of \emph{forward bias}. 
This is the apparent reason why the Reverse Recall task fails.

\subsection{Transfer learning of encoder pretrained on Reverse Recall}

\begin{figure}[!h]
\centerline{\includegraphics[width=0.7\columnwidth]{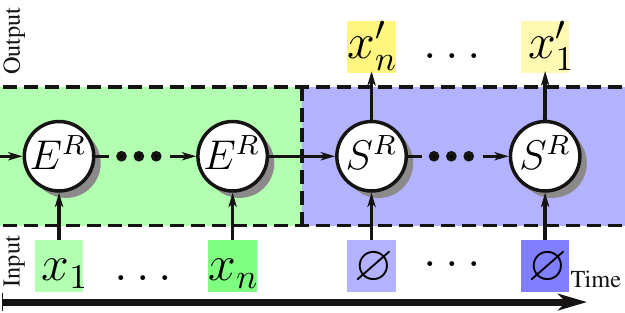}}
\caption{MAES model trained on Reverse Recall task in an end-to-end manner}
\label{fig:reverse_recall_err_srr}
\end{figure}

To completely eliminate the forward bias so that each element is encoded independent of the others,
we considered a new encoder-solver model trained on Reverse Recall task from the scratch in an end-to-end manner (\fig{reverse_recall_err_srr}).
So the role of encoder $E^R$ (from Encoder-Reverse) is to encode and store inputs in the memory, wherein the solver $S^R$ is supposed to produce the reverse of the sequence.
Because we do not allow unbounded jumps in attention in the design of MAES, we add an additional step
where in the solver design we also initialize the read attention
of the solver to be the write attention of the encoder at the end of processing the input. This way the
solver can possibly recover the input sequence in reverse by learning to shift the attention in the reverse order.

Generally our hypothesis was that the encoder trained by this process should be free of \textit{forward bias},
via the following reasoning. 
Consider what happens if we learned a perfect encoder-solver for producing the reverse of the input
for sequences of all lengths.
Let the input sequence be $x_1, x_2,\dots, x_n$ for some arbitrary $n$ where $n$ is not known to
the encoder in advance.
Assume that similar to the earlier case of encoder $E^S$, we have encoded this sequence as
$z_1, z_2, \dots, z_n$ where for each $k$,
$z_k = f_k(x_1, x_2, \dots, x_k)$ for some function $f_k$.
To have no forward bias, we need to show that $z_k$ depends only on $x_k$, i.e. $z_k = f_k(x_k)$.
Then for the hypothetical sequence $x_1, x_2, \dots, x_k$ the encoding of $x_k$ will still equal
$z_k$ since the length of the sequence is not known in advance.
For this hypothetical sequence, the solver starts by reading $z_k$.
Since it has to output $x_k$, the only way for this to happen is when there is one-to-one mapping
between the set of $x_k$'s and the set of $z_k$'s. Thus $f_k$ depends only on $x_k$ and there is no
forward bias.
Since $k$ was chosen arbitrarily, this claim holds for all $k$, showing that the resulting encoder should have 
no forward bias.

While the above approach was promising, it hinged on the assumption of perfect learning and it's not clear what happens if 
that is not the case.
In our experiments we were able to achieve validation accuracy of 100\% for decoding the forward as well 
as reverse order of the input sequence (i.e. Serial and Reverse Recall tasks).
But the training did not converge and the best loss function value was about 0.01.
With such a large training loss, the task-size generalization worked well 
for sequences up to length 500, achieving  perfect 100\% accuracy (with a large enough memory size).
However, beyond that length, the performance started to degrade, and at length 1000, the test accuracy
was only as high as 92\%. 
This result motivated the design of better encoders that work universally for both forward and reverse oriented tasks.

\subsection{Improved encoder using Multi-Task Learning}

\begin{figure}[!h]
\centerline{\includegraphics[width=0.7\columnwidth]{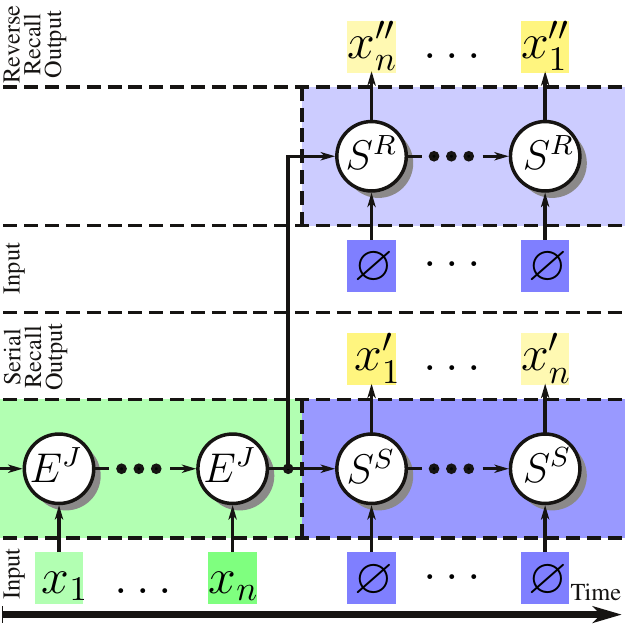}}
\caption{MAES model used for joint training of Serial and Reverse Recall tasks}
\label{fig:recall_ej_srr_ssr}
\end{figure}

To obtain an improved encoder capable of handling both forward and reverse-oriented sequential tasks,
 we use a Multi-Task Learning (MTL)~\cite{caruana1993multitask} approach using hard parameter sharing, 
i.e. build a model having a single encoder and many solvers.
However, the key difference is that we do not jointly train on all the tasks.
In the model presented in \Fig{recall_ej_srr_ssr}, we explicitly enforce the encoder ($E^J$ from Encoder-Joint) to produce an encoding that is simultaneously good for both the Serial and Reverse Recall tasks. The goal is to see
whether this form of \emph{inductive bias} that was used in the design of the encoder is also
sufficient to build good solvers independently for other sequential tasks.

\begin{figure}[!t]
\centerline{\includegraphics[width=0.87\columnwidth]{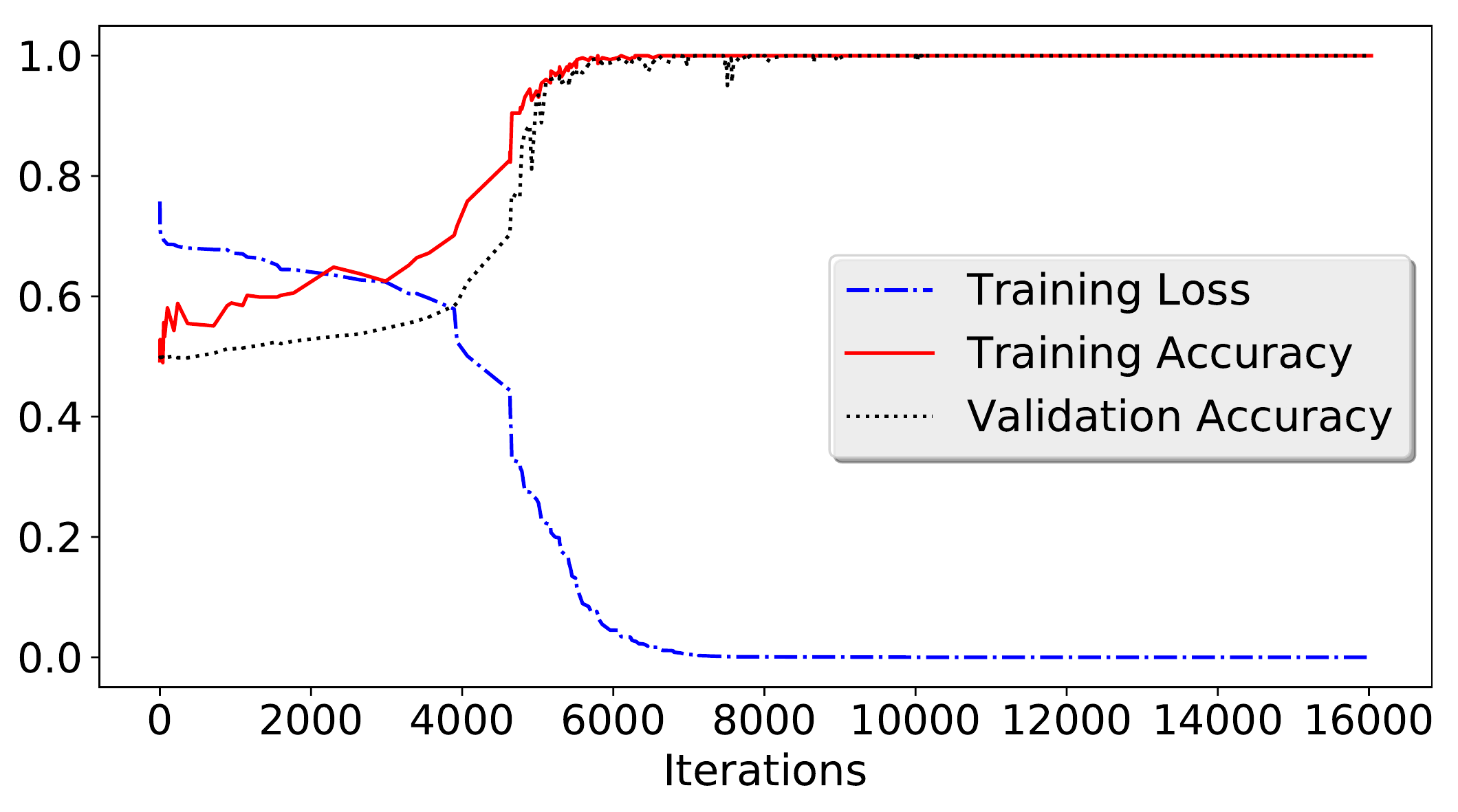}}
\caption{Training performance MAES model trained jointly on Serial Recall and Reverse Recall tasks.
Training loss of $10^{-5}$ obtained after \texttildelow{}12,000 iterations}
\label{fig:gp-enc-training}
\end{figure}

\Fig{gp-enc-training} shows the training performance of building the encoder via this approach. 
Compared to the training of the first encoder $E^S$ the training loss takes a longer time to start dropping
but still the overall convergence was only about 1000 iterations longer compared to the encoder $E^S$.
However, as presented in \Fig{gp-enc-same-input}, we have the desired property that 
the encoding of the repeated sequence stored in memory is near-uniform across all the locations, 
demonstrating the elimination of forward bias.

\begin{figure}[!b]
    \centering
    \includegraphics[width=0.9\columnwidth]{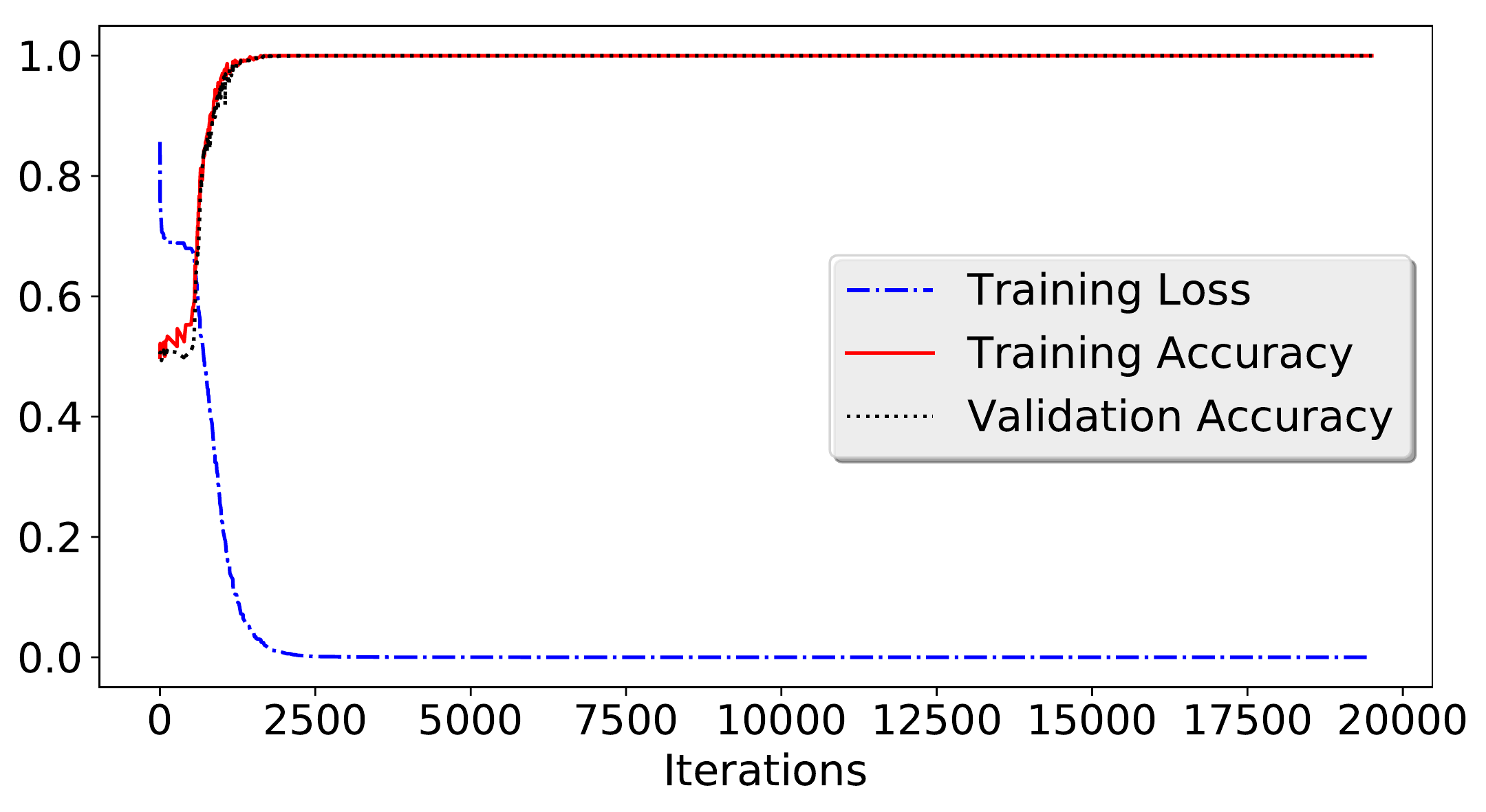}
    \caption{Sequence Comparison Task}
     \label{fig:seq-comp-training}
\end{figure}\

\begin{figure}[!t]
    \centering
    \includegraphics[width=0.9\columnwidth]{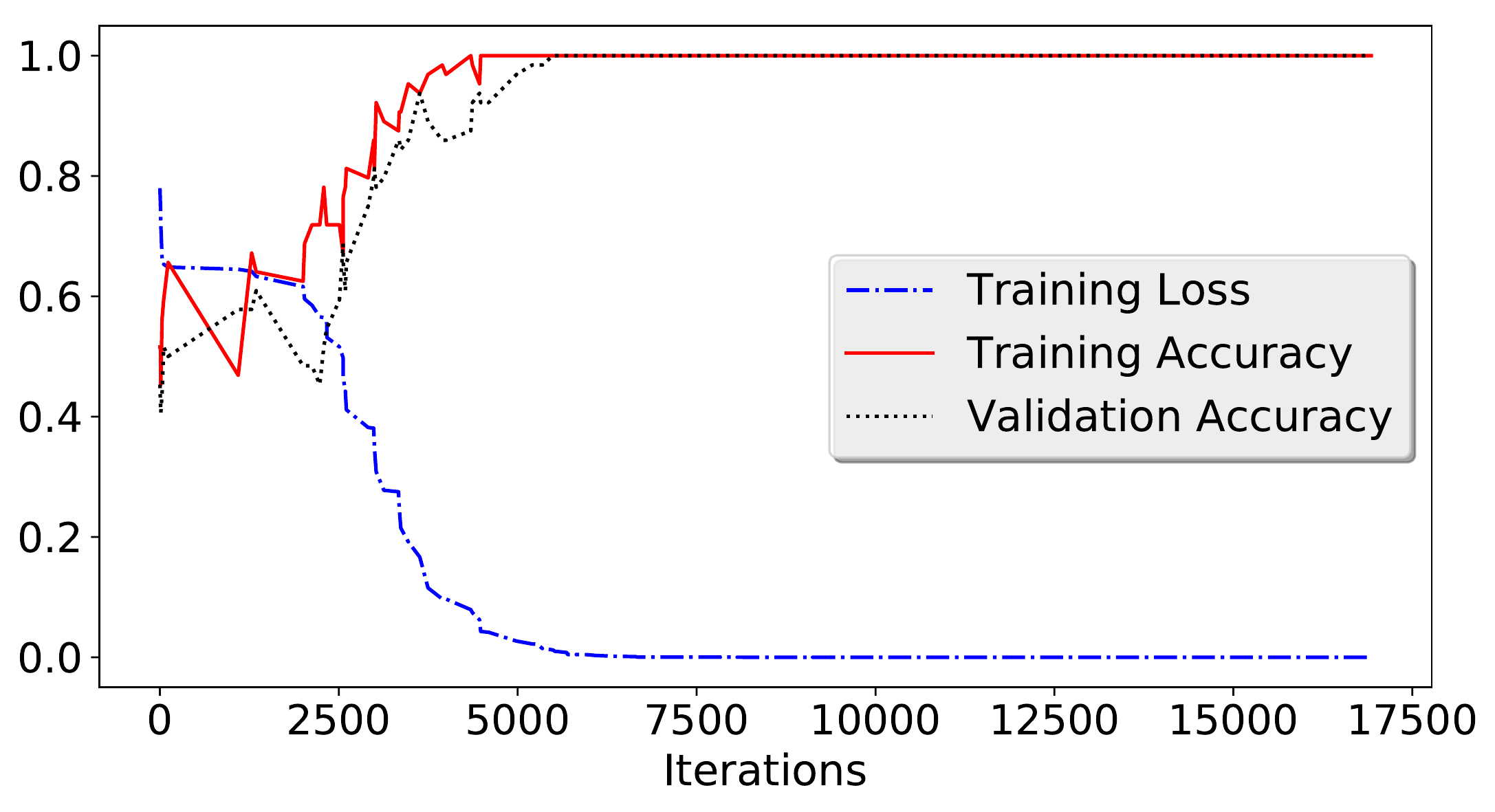}
    \caption{Equality Task}
     \label{fig:equality-training}
\end{figure}

\section{Performance of MAES with pretrained joint encoder}
Below we describe the results for the various working memory tasks using this the encoder $E^J$.
In all of these tasks the encoder $E^J$ was frozen and we trained only task-specific solvers.
The aggregated results can be found in \Tab{ej-results}.

\subsection{Serial Recall and Reverse Recall}
Since the encoder $E^J$ was designed with the purposes of being able to do both tasks well (depending on where the attention is given to the solver), it's not surprising that we should do better than training them end-to-end individually.
Indeed the training for Reverse Recall is quite fast and for Serial Recall it compares more favorably compared to the encoder $E^S$.

\subsection{Odd Task}
For this task, we started with the $E^J$ encoder and a solver which has only the basic attention shift mechanism i.e.
able to shift at most 1 memory address in each step. We wanted to verify that this should not train well since the attention 
on the encoding needs to jump by 2 locations in each step. Indeed the training did not converge at all with a loss value close
to 0.5.
After adding the additional capability for the solver to be capable of shifting its attention by 2 steps, 
the model converged in about 7,200 iterations.


\subsection{Sequence Comparison and Equality Tasks}
\label{sec:seq-eq}
Both these tasks involve comparing the solver's input element-wise to that of the encoder, so to 
compare their training performance, we used the same parameters for both the tasks.
In particular, it resulted in the largest number of trainable parameters due to the additional hidden layer 
(with ReLU activation). 
Since Equality is just a binary classification problem, having small batch sizes 
caused enormous fluctuations in the loss function during training.
Choosing a larger batch size of 64 stabilized this behavior and  allowed the training to converge in about 11,000 iterations for
Sequence Comparison (\Fig{seq-comp-training}) and about 9,200 iterations for 
Equality (\Fig{equality-training}).
While the wall time was not affected by this larger batch size (due to efficient GPU utilization), 
it is important to note that the number of data samples is indeed much larger than that for the
other tasks.

Equality exhibits larger fluctuations in the initial phase of training due to the
loss being averaged only over 64 values in the batch. But it also converged faster which seems
counterintuitive because the information available to the trainer is just a single bit for the Equality task.
We believe that this happened because the distribution of instances to the Equality problem is such
that even with a small number of mistakes on the individual comparisons there exists an error-free
decision boundary for separating the binary classes. We conjecture that this always the case for any 
distribution of instances for Equality.

\begin{table*}[t]
\vskip 0.15in
\begin{center}
\begin{small}
\begin{sc}
\begin{tabular}{lcccccccc}
\toprule
Task & \multicolumn{2}{c}{MAES } & \multicolumn{2}{c}{LSTM} & \multicolumn{2}{c}{NTM} & \multicolumn{2}{c}{ DNC}\\
\multirow{2}{*}{}& Steps to & Task-size & Steps to & Task-size & Steps to & Task-size & Steps to & Task-size \\ 
& converge & general. & converge & general. & converge & general. & converge & general. \\
\midrule
Serial Recall & 7,000 & $100\%$ & 100,000 & $50.12\%$ 	& 29,000 & $100\%$ &  19,100 & $65.64\%$ \\
Reverse Recall & 6,900 & $100\%$ & 100,000 & $50.19\%$ 	& 37,000 & $100\%$ & 17,800 & $63.34\%$ \\
Odd & 7,200 & $100\%$ & 100,000 & $50.72\%$ 				& 57,000 & $100\%$ &  100,000 & $50.19\%$ \\
Equality & 9,200 & $100\%$ & 100,000 & $50.13\%$ 		& 98,000 & $100\%$ & 83,100 & $45.31\%$ \\
Comparison & 11,000 & $100\%$ & 100,000 & $50.75\%$  	& 100,000 & $50.13\%$ & 100,000 & $50.75\%$ \\
\bottomrule
\end{tabular}
\end{sc}
\end{small}
\end{center}
\vskip -0.1in
\caption{Results for MAES using encoder $E^J$ (i.e. pretrained jointly on Serial and Reverse Recall tasks) trained on the set of tasks along with the results achieved by the baseline models}
\label{tab:ej-results}
\end{table*}

\subsection{Baseline Results}
In order to compare the achieved results we have implemented three baseline architectures.
As first model we used stacked LSTM~\cite{hochreiter1997long} with 3-layers and 512 hidden units in each of them.
We decided to provide results for this model as it is treated as a standard reference baseline for all recurrent neural models.
Besides, we provide baseline calculations with two state-of-the art Memory-Augmented Neural Networks: Neural Turing Machine (NTM)~\cite{graves14} and Differentiable Neural Computer (DNC)~\cite{graves2016hybrid}.
The NTM had a simple RNN controller with sigmoid activation function and 20 hidden units, whereas the DNC incorporated an LSTM controller with a single layer with 20 hidden units. The calculations were terminated if the loss dropped below 1e-5 or the calculation reached 100,000 episodes. 
To obtain optimal convergence, the baseline results used curriculum learning up to sequences of length 20 with early stopping judged on validation sequences of length 21. 
For each baseline architecture and task pair we have trained 10 models and picked the best one taking into consideration the smallest number of episodes required by a given model to converge.
Those models were used during testing of task-size generalization (i.e. on sequences of length 1000).
For some tasks some models didn't converge at all i.e. calculation reached 100,000 episodes and validation loss was still above the threshold.
In those cases we report the results for models that had the smallest validation loss.
The results are presented in~\Cref{tab:ej-results}.

In agreement with the results presented in \cite{graves14} the LSTM can be trained on simple serial recall for sequences up to length 20 but the loss does not converge to our very tight threshold. This lack of convergence extended to all of the tasks and as expected the LSTM was incapable of generalizing to sequences of length 1000. As~\Cref{tab:ej-results} shows, the LSTM did no better than random guessing implying that it did not learn to solve the task for all input sizes. This was expected as the LSTM's hidden dimension must grow proportionally to the size of the task's input.

NTM results are much better, i.e. it managed to converge and achieved task-size generalization in four out of five tasks.
However, it is important to note that for Equality actually only one model managed to master the task.
Aside of that, it is worth noting that, in comparison to MAES, NTM models required much longer training.

Finally, the DNC obtained a loss of better than 1e-3 on all tasks, even though it did not obtain the strict convergence threshold of 1e-5. While it learned the tasks for short sequences, in all cases the DNC struggled with generalization to sequences of length 1000. Thus, it did not achieve task-size generalization, but our additional tests did show that the DNC could generalize to sequences of length 100 on serial recall. We hypothesize that the DNC, which was designed for episodic memory tasks, had difficultly generalizing on strict sequential tasks, because the relative complexity of the read and write mechanisms raised the probability of making mistakes on extremely long sequences.




%% file: summary.tex
\section{Summary and future Directions}
\label{sec:summary}
In this paper we have introduced a novel neural architecture called Memory-Augmented Encoder-Solver.
It is derived from the encoder-decoder architecture and additionally contains a block of memory that is retained between encoder and solver during task execution.
The modularity of the proposed architecture enables both transfer and multi-task learning.
According to our knowledge this is the first application of those concepts in the area of Memory-Augmented Neural Networks.
The resulting models are extremely fast to train and their performance clearly beats the performance of other state-of-the-art models.

An important class of working memory tasks not considered here but worthy of investigation are 
``memory distraction'' tasks. The characteristic of such dual tasks is the ability to handle
multiple interruptions in the middle of solving a main task in order to solve temporary tasks.
This is challenging to solve in our MAES framework because it requires suspending encoding of the main input arbitrarily 
in the middle to solve the distraction task and finally return attention to where the encoder was suspended. 
Handling this in the MAES framework (with or without pretrained encoders) requires new ideas that we are currently exploring.
Another challenge is to handle some of the popular visual working memory tasks in psychology that require finding suitable encodings for images. In principle the same overall design considered in this paper could also extend to such tasks, however, whether the learning can be accomplished in such a modular fashion remains an interesting open question. From the standpoint of applications, recently the encoder-decoder framework along with attention models have been used to improve the performance of machine translation problems~\cite{bahdanau2014neural, vaswani2017attention}. Incorporating these models into the MAES framework will require content-addressable memories, which are not part of working memory, but leave open the possibility of designing hybrid models to tackle such problems more generally in the area of language modeling.
Finally, whether we can use the multi-task transfer learning on working memory tasks to help train multi-modal problems like question answering or visual question answering is an interesting open question.